# EOE: Expected Overlap Estimation over Unstructured Point Cloud Data

Ben Eckart   Kihwan Kim   Jan Kautz

NVIDIA Research

{beckart, kihwank, jkautz}@nvidia.com

http://research.nvidia.com/publication/2018-09_Probabilistic-Overlap-Estimation

## Abstract

*We present an iterative overlap estimation technique to augment existing point cloud registration algorithms that can achieve high performance in difficult real-world situations where large pose displacement and non-overlapping geometry would otherwise cause traditional methods to fail. Our approach estimates overlapping regions through an iterative Expectation Maximization procedure that encodes the sensor field-of-view into the registration process. The proposed technique, Expected Overlap Estimation (EOE), is derived from the observation that differences in field-of-view violate the* iid *assumption implicitly held by all maximum likelihood based registration techniques. We demonstrate how our approach can augment many popular registration methods with minimal computational overhead. Through experimentation on both synthetic and real-world datasets, we find that adding an explicit overlap estimation step can aid robust outlier handling and increase the accuracy of both ICP-based and GMM-based registration methods, especially in large unstructured domains and where the amount of overlap between point clouds is very small.*

## 1. Introduction

Point cloud registration recovers the spatial transformation between two or more point clouds by matching and aligning their common geometry. Most classic registration algorithms are framed as an iterative two-step process that cycles between data association (matching) and minimizing the distance between the matched data (aligning).

In many real-world scenarios where range sensors collect data at varying viewpoints (e.g. LiDAR on a moving vehicle), limited field-of-view can produce large amounts of non-overlapping point data. This may cause problems for the matching step of registration algorithms since non-overlapping data has no valid data association. Traditional registration methods over unstructured data generally rely on two strategies to combat these problems: 1) some form of outlier detection to remove unmatched data during association [9, 26, 23], or 2) robust optimization to prevent spuriously matched data from corrupting the solution [5, 4, 21, 13]. In this work we propose a different approach: an estimation procedure that attempts to preserve the underlying data distribution assumptions made by the registration algorithm through field-of-view constraints.

Our approach is derived from the observation that differences in field-of-view violate the *iid* assumption implicitly held by nearly all registration algorithms. That is, most of these methods are based on the following problematic assumption: *That there exists some latent probability distribution for which point cloud data can be viewed as a set of independent and identically (iid) distributed samples.*

In this paper, we explore how violations of this basic statistical assumption directly lead to various failure cases when utilizing popular registration techniques on real world data. These failure cases motivate us to propose a new augmentation to existing registration techniques, Expected Overlap Estimation (EOE), which attempts to mitigate these violations in real-world scenarios. We concentrate on overlap estimation due to the observation that differences in field-of-view under changing viewing angles are often a primary cause of poor performance in real-world scenarios. Given the additional ability to estimate what has only been seen once between two views, the registration problem (including basic outlier detection) becomes much more tractable and accuracy often increases significantly. See Figure 1 for a depiction of this idea.

Our contributions are summarized as follows: We establish a view-aware methodology for data association that explicitly considers only overlapped regions of geometry (Sec. 3.1). We then provide an iterative procedure to dynamically estimate view overlap during registration, which we call Expected Overlap Estimation (EOE) (Sec. 3.2). We deploy this method as an augmentation to several state-of-the-art registration algorithms, both GMM-based and ICP-based, and show clear improvements in robustness on both synthetic and real-world data, especially in cases of non-overlapping fields-of-view.

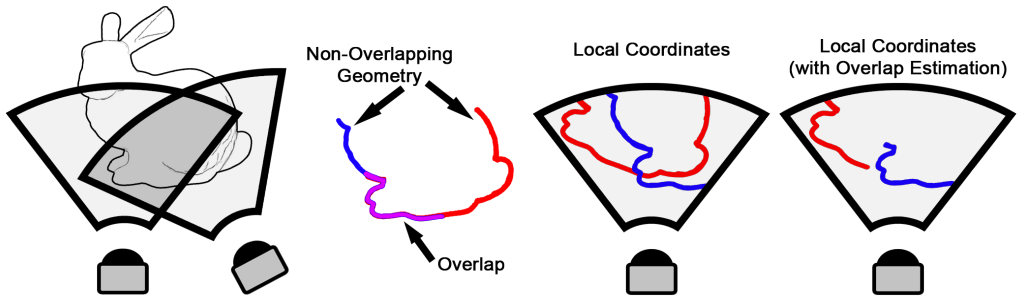

Figure 1. **Point Cloud Registration with Limited Overlap** Leftmost: Two range sensors at different poses capture different slices of geometry. Note that the geometry is illustrated in 2D contours for simpler visualization: blue and red contours for the geometry captured from each view, and magenta for overlapped region. During the registration process, the non-overlapping pieces of geometry are problematic as they may be mistakenly matched. For most registration methods, this situation manifests itself as a violation of the fundamental assumption that both point sets are *iid* samples of a single distribution. Rightmost: If overlap can be directly estimated as part of the model, then non-overlapping regions could be downweighted or ignored during data association and matching, leading to a more well-formed registration problem both practically and theoretically.

## 2. Related Work

Ever since the Iterative Closest Point (ICP) algorithm was introduced as a general method for registering 3D shapes over two decades ago [1, 3], the topic of 3D point cloud registration has remained a challenging and core problem in the field of 3D perception, especially in recent years given the current growing interest in virtual and mixed reality [19] and autonomous vehicles [12, 18].

The introduction of ICP has since spawned many variants and augmentations, which can be generally divided into two types of algorithms: 1) attempts to improve on specific subtasks (e.g. selecting, matching, or aligning [23]) while adhering to ICP's original structure or 2) statistical generalizations of the ICP algorithm in order to leverage more advanced models (e.g. Gaussian Mixture Models [17, 16, 7, 8]) and algorithms like Expectation Maximization [11, 14]. We will refer to the former type as "ICP-based" algorithms and the latter type as "GMM-based" algorithms.

Our proposed technique of Expected Overlap Estimation shares similar motivations to previous work on robust outlier detection and handling under ICP-based algorithms. For example, FICP [21], IRLS-ICP [13], and Trimmed ICP (TrICP) [5, 4] utilize robust matching metrics in order to handle cases of partial overlap in order to down-weight and/or avoid spurious point matches. In particular, FICP (Fractional RMSD ICP) works by only considering a fraction of the closest matches inside the distance minimization (RMSD) term. Thus, the worst fraction of matches (defined as a parameter $f \in [0, 1]$) will be ignored, with the intent of causing the distance minimization to only include points with valid matches.

We similarly seek to detect points that have no geometry with which to match in the other point cloud, though we use an iterative approach where field-of-view estimated are dynamically updated and used to constraint both data association and match optimization. We will show that both approaches are complementary to one another.

Another related approach to exclude non-overlapping regions is to use projective association for the matching step of ICP [23]. Projective ICP has seen a lot of success recently as part of a family of "dense" techniques. For example, KinectFusion [20], performs projective association by utilizing the dense 2D depth maps from the Kinect. Projective association can be seen as a type of implicit ray casting solution to remove data currently believed to be outside the region of overlap. This efficient method first transforms a pair of 2D depth maps into a single coordinate system and then pairs depth pixels sharing the same 2D index. However, this type of data association relies on a dense 2D grid of depth points, and it is not clear how one would ray cast through sparser data, such as from a Lidar, to associate points.

Our proposed method can be seen as a probabilistic generalization of the "hard assignment" scheme of projective association: instead of ray casting to find a hard data association, we utilize the machinery of Expectation Maximization to obtain "soft labels" to characterize our projective association given some underlying spatial probability model. In contrast to normal projective association where the hard association is used simply for pair matching, our soft labels serve two purposes: 1) to downweight points with no likely projective association to the spatial point model, and 2) to modify the underlying spatial model to discount areas outside the estimated area of overlap. This probabilistic framework therefore both generalizes projective methods and also allows them to be used with unstructured point cloud data (e.g. Lidar data) and with any underlying probability model, making it useful for both ICP-based and GMM-based registration algorithms.

## 3. Approach

The basic Maximum Likelihood Estimate (MLE) criterion used for most modern registration methods assumes both point sets are *iid* samples of a single generative model, and that one point set has undergone some latent spatial transformation. Given two point sets $\mathcal{Z}_1$ and $\mathcal{Z}_2$, the MLE transformation $T$ therefore maximizes the data probability of the transformed point cloud $T(\mathcal{Z}_2)$ given some probabilistic representation of spatial likelihood derived from the spatial distribution of the first point cloud $\mathcal{Z}_1$. If we parameterize our probability model by $\hat{\Theta}$, we obtain the following formulation:

$$\hat{T} = \underset{T}{\operatorname{argmax}}\, p(T(\mathcal{Z}_2)|\hat{\Theta}_{\mathcal{Z}_1}) \quad (1)$$

Since the points are assumed to be *iid* samples of $\hat{\Theta}$, the data likelihood further decomposes into a product over all points in $\mathcal{Z}_2 = \{\mathbf{z}_i\}, i = 1..N_2$,

$$\hat{T} = \underset{T}{\operatorname{argmax}} \prod_{i}^{N_2} p(T(\mathbf{z}_i)|\hat{\Theta}_{\mathcal{Z}_1}) \quad (2)$$

Whether one or both point sets were used in the creation of $\Theta$, however, there is the implicit assumption that both $\mathcal{Z}_1$ and $\mathcal{Z}_2$ represent *iid* samplings of the same probabilistic distribution. By adopting this framework, we assume that there is a latent probabilistic function controlling the generation of points, the specific form of $p(z|\hat{\Theta})$ dictating the individual point probabilities of 3D space.

### 3.1. Overlap Estimation Given Known Views

Points from geometry present in *non-overlapping view regions* of each point set make the common construction of most registration algorithms ill-posed. Since a probability distribution describing each point set will have fundamentally different spatial extents, these view differences make it impossible to represent both point sets as the *iid* sampling of the same latent model. This limits current registration algorithms' applicability in cases where large transformations, limited field-of-view, or occlusions between point sets cause large degrees of non-overlapping geometry.

See Fig 2 for a depiction of this violation. Let $A$ and $B$ be the views from which $\mathcal{Z}_1$ and $\mathcal{Z}_2$ are captured, respectively. Under the traditional maximum data likelihood paradigm, interpreting $\mathcal{Z}_2$ as a set of rigidly transformed *iid* samples of the model constructed from $\mathcal{Z}_1$ ($\Theta_{\mathcal{Z}_1}$) is problematic since $\Theta_{\mathcal{Z}_1}$ only "knows" about the space within region $A$. If $B \cap \neg A \neq \emptyset$, then any $\mathbf{z}_i \in \mathcal{Z}_2$ inside this region will be unexplainable by $(\Theta_{\mathcal{Z}_1})$. Similarly, if $A \cap \neg B \neq \emptyset$ and $\int_{A \cap \neg B} p(z|\Theta_{\mathcal{Z}_1})dz > 0$ then $\mathcal{Z}_2$ will become more and more unlikely to be considered an *iid* sampling from the model described by $\Theta_{\mathcal{Z}_1}$ as the cardinality of $\mathcal{Z}_2$ increases, since there will always be a portion of the model ($A \cap \neg B$) from which samples never appear. Even in the case of a symmetric global model where $\Theta$ is derived from both $\mathcal{Z}_1$ and $\mathcal{Z}_2$, as in JRMPC [8], it is still the case that no single point cloud can be considered an *iid* sampling of the global model if $A \neq B$, and thus non-overlapping point data may be erroneously matched during registration.

To tackle these problems, we start with a goal of being able to incorporate and leverage known sensor view constraints into the registration process. Given an estimated field-of-view, $\Psi$, of the sensor (i.e. minimum range $\Psi_{min}$ and maximum range $\Psi_{max}$ distances and horizontal and vertical field-of-view angles $\Psi_x$, $\Psi_y$), which can often be found by reading the sensor's data sheet, if we knew the relative pose change between A and B, we could also calculate where the overlapping regions occur. Figure 1 illustrates this idea using a partially non-overlapping Stanford Bunny. Denoting $\Omega$ as our overlapping region (Fig. 2), we can then augment the registration problem as follows:

$$\hat{T} = \underset{T}{\operatorname{argmax}}\, \Omega^*(T(\mathcal{Z}_2)|\Theta_{\mathcal{Z}_1}), \forall \mathcal{Z}_2 \in \Omega \quad (3)$$

Where $\Omega^*(\cdot)$ denotes the functional,

$$\Omega^*(p) = \begin{cases} \eta p(z), & \text{if } z \in \Omega \\ 0, & \text{otherwise} \end{cases} \quad (4)$$

where $\eta$ is a normalization constant to enforce $p(z)$ summing to 1 over all $\Omega$.

This formulation better conforms to the *iid* assumption underlying statistical point cloud models in that it defines registration to occur only over previously seen regions of space. Equation 3 maximizes the data probability only over $\Omega$: the probabilistic model is modified to only reflect the distribution of $\mathcal{Z}_1 \in \Omega$, and the data likelihood only considers samples from $\mathcal{Z}_2 \in \Omega$. We therefore avoid some of the *iid* violations resulting from the typical maximum likelihood construction, where points in $\mathcal{Z}_2$ not in $\Omega$ appear in the data likelihood product, and where $\Theta_{\mathcal{Z}_1}$ is partly derived from regions not in $\Omega$.

### 3.2. Expected Overlap Estimation

Of course, if we knew the overlapping region $\Omega$ between two views A and B (and thus the relative pose change between A and B), we would already have exactly what we are trying to find when we perform registration.

This formulation does, however, present us with an iterative approach for imposing a general view-awareness on any iterative registration algorithm. That is, if we impose a probability function over the estimation of $\Omega$ using known sensor view properties $\Psi$ and relative transformation $T$, we can iteratively hold $\Omega$ fixed to estimate $T$ and then hold $T$ fixed to re-estimate $\Omega$ (and also therefore the functional $\Omega^*$). This construction forms an Expectation Maximization procedure: In our E-Step, we calculate the expected latent model of overlap ($\Omega$) with respect to a posterior over our data while holding $T$ fixed. In the M-Step, we find the maximum likelihood $T$ using our expectation of $\Omega$ from the previous E-Step. More rigorous mathematical details on this construction can be found in the supplementary materials.

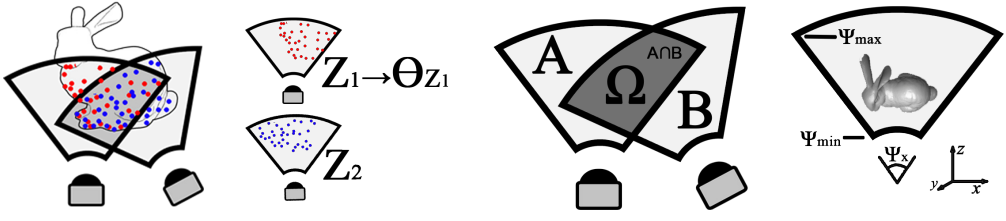

Figure 2. **Definitions** *Left:* A range sensor from two different viewpoints produces two point clouds $\mathcal{Z}_1$ and $\mathcal{Z}_2$. A model from $\mathcal{Z}_1$ is constructed (parameterized by $\mathbf{\Theta}_{\mathcal{Z}_1}$), either explicitly (for GMM-based approaches) or implicitly (through least squares minimization, as with ICP). *Right:* We denote the respective field-of-view regions as A and B and overlapped region as $\Omega$. We group together known sensor parameters related to the field-of-view as $\Psi = \{\Psi_{min}, \Psi_{max}, \Psi_x, \Psi_y\}$. $\Psi_{min}$ and $\Psi_{max}$ denote the minimum and maximum range of the sensor, $\Psi_x$ denotes the viewing angle along the x-axis in radians, and $\Psi_y$ denotes the viewing angle along the y-axis in radians (not shown in this top-down 2D depiction).

**Algorithm 1** Expected Overlap Estimation (EOE)

**Step 1:** $\hat{\mathbf{\Theta}}_{\mathcal{Z}_1} = \underset{\mathbf{\Theta}}{\mathrm{argmax}}\, p(\mathcal{Z}_1 | \mathbf{\Theta})$
**Step 2:** $\hat{T} = \underset{T}{\mathrm{argmax}}\, \hat{\Omega}^*(T(\mathcal{Z}_2) | \hat{\mathbf{\Theta}}_{\mathcal{Z}_1}), \forall Z_2 \in \hat{\Omega}$
**Step 3:** $\hat{\Omega} = \underset{\Omega}{\mathrm{argmax}}\, p(\Omega | \hat{T}, \Psi)$
**Step 4:** If not converged, go to Step 2.

**Algorithm 2** $\Omega$-Estimation

1: **procedure** CALC_$\Omega(\mathcal{Z}$, R, t)
2:    **camera constants:** $\Psi_x, \Psi_y, \Psi_{min}, \Psi_{max}$
3:    **penalty constants:** $k_0, k_1, k_2$
4:    z_weights $\leftarrow \vec{1}$ // $N \times 1$ array of weights for each $\mathbf{z}_i \in \mathcal{Z}$
5:    **for** $\mathbf{z}_i \in \mathcal{Z}_2$ **in parallel do**
6:       $\xi \leftarrow 0$ // initialize per-point penalty term
7:       $\tilde{\mathbf{z}}_i \leftarrow R^T(\mathbf{z}_i - t)$ // project $\mathcal{Z}$ into common frame
8:       $d \leftarrow \|\tilde{\mathbf{z}}_i\|$ // distance of point from other sensor
9:       $\theta \leftarrow \mathrm{atan2}(\tilde{\mathbf{z}}_{iy}, \tilde{\mathbf{z}}_{ix})$ // horiz. angle of $\mathbf{z}_i$ rel. to $\Psi_x$
10:      $\phi \leftarrow \mathrm{acos}(\tilde{\mathbf{z}}_{iz}/d)$ // vert. angle of $\mathbf{z}_i$ rel. to $\Psi_y$
11:      // check range bounds
12:      **if** $\{d < \Psi_{min}$ **or** $d > \Psi_{max}\}$ **then** $\xi \leftarrow \xi + k_0$
13:      // check angular bounds on $\Psi_x$
14:      **if** $\{\theta > \Psi_x/2$ **and** $\theta < 2\pi - \Psi_x/2\}$ **then**
15:         $\xi \leftarrow \xi + \min(\theta - \Psi_x/2.0, 2\pi - \Psi_x/2.0 - \theta)$
16:      **end if**
17:      // bounds on $\Psi_y$
18:      **if** $\{2\phi > \pi + \Psi_y\}$ **then** $\xi \leftarrow \xi + \phi - (\pi/2 + \Psi_y/2)$
19:      **if** $\{2\phi < \pi - \Psi_y\}$ **then** $\xi \leftarrow \xi - \phi + (\pi/2 + \Psi_y/2)$
20:      // downweight if FoV violation
21:      **if** $\{\xi > 0\}$ **then** z_weights[i] $\leftarrow k_1 e^{-k_2 \xi}$
22:    **end for**
23:    **return** z_weights
24: **end procedure**

Algorithm 1 shows each general step of EOE algorithm. Note that we can use any standard registration algorithm for the first two steps, making this a general and complementary augmentation for most existing registration algorithms. We will now describe each step in more detail.

**Step 1**: For ICP-based methods, we can effectively skip this step since the model is built implicitly into the distance minimization. If using an NDT construction [2, 24, 25], we simply voxelize 3D space and then calculate local voxel point distributions. For other general GMM-based techniques such as MLMD, ECMPR, or JRMPC for example, we use the EM algorithm [7, 8, 14].

**Step 2**: The basic optimization underlying Step 2 has been solved in many different ways, each with their own tradeoffs. Many methods rely on simplifications to the structure of $\mathbf{\Theta}$, including: imposing the same covariance to every mixture, restricting covariances to be isotropic, or equal mixture weighting so that a closed form solution may be applied (typically Horn's method [15]). ICP, EM-ICP, CPD, and JRMPC can be considered examples of this. Other approaches put less restrictions on the structure of $\mathbf{\Theta}$, but using gradient descent [6] in lieu of a closed form solution. Others have approximated the optimization criterion itself to a more tractable form, for example, by using direct simplification [7] or SDP relaxation [14]. To provide field-of-view awareness, we use the current best guess for $\hat{\Omega}$ to apply $\Omega^*(p)$, using the previous weights from the last application of Step 3. For ICP-based methods, this can be done by weighting the points using the expectations from Step 3. For GMM-based methods, we additionally re-weight the model by applying Equation 4 using our previous guess of $\hat{\Omega}*$.

**Step 3**: For Step 3, we need to establish a probability model to reason over potential choices of $\Omega$. In this paper, we adopt an exponential likelihood field approach. Using our current estimate of $\hat{T} = \{R, t\}$, we alternately project points into each sensor's local field-of-view defined by $\Psi$. Projected points that fall outside each local field-of-view estimate are downweighted expontially according to the user-defined constants $k_0, k_1, k_2$. This likelihood field serves as a heuristic for an expected value calculation over a latent indicator function for points inside $\Omega$. Refer to Algorithm 2 for more details.

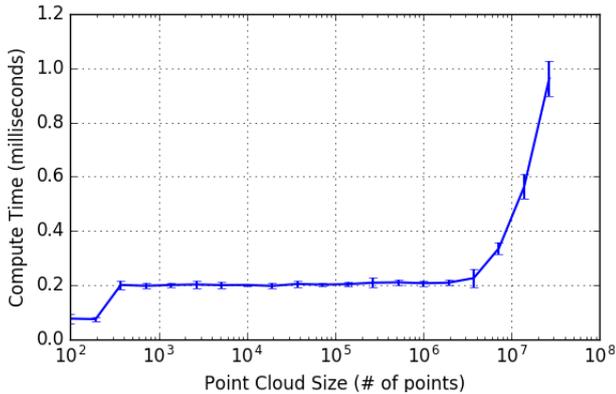

Figure 3. **Computational Scaling of Algorithm** 2 We implemented Algorithm 2 using CUDA, and tested it on an NVIDIA Titan X GPU. Compute time (in milliseconds) is plotted against the number of points in the point cloud for increasing point cloud sizes. Even for inputs up to millions of points, the compute time remains under 1 ms.

*Step 4*: Convergence is defined by a set number of max iterations, but with an early stopping condition if $\Delta T$ falls below a threshold.

## 4. Evaluation

### 4.1. Computational Efficiency

As shown in Algorithm 2, by representing $\Omega$ as a likelihood field over the intersection of the field-of-view extents for each sensor, the additional operations Steps 2 and 3 in Algorithm 1 can be done very efficiently and in parallel, because each point requires only a single constant-time boundary check and reweighting per iteration and this computation can be done independently per point. Given that each point can be checked in parallel, Algorithm 2 becomes a natural fit for GPU acceleration, which we have done by implementing it in CUDA as a single kernel. In some cases, the overhead of these checks can be offset by the speed up of the base algorithm, since discounting points outside $\Omega$ result in a smaller sized or more easily convergent registration problem.

Figure 3 shows the computational speed of our CUDA-based implementation of Algorithm 2 when using an NVIDIA Titan X GPU. On the x-axis, we vary the input point cloud size from 100 to over 1 million points, and on the y-axis we record the time to load and compute a new set of weights given a random transformation. We can see that for most point cloud sizes between 1000 and 1 million, EOE adds about 0.2 ms overhead per iteration. Given a typical ICP algorithm running for 20 iterations, EOE augmentation would add roughly 4 ms to the total running time, a fairly negligible amount relative to the speeds of most current registration algorithms.

### 4.2. Experimental Set-Up

We compare experimental results with and without EOE over various datasets and various algorithms: ICP [3], MLMD [7], Trimmed ICP [5], Fractional ICP [21], and Iteratively Reweighted Least Squares ICP [13]. We choose ICP since it is still widely used in this domain, MLMD because it represents a recent GMM-based method, and TrICP, FICP, and IRLS-ICP because they are commonly used open source methods and employ robust outlier handling. For each method, and for the rest of our experiments, we used the default parameters suggested by the authors, or in the case of TrICP, FICP, and IRLS-ICP, from the sample configuration files in the open source *libpointmatcher* package [22]. Note that given our focus on unstructured and/or sparse point cloud data, we cannot test against methods like projective ICP that require dense 2D depth maps [23]. Our testbed has an Intel i7-5820K CPU at 3.30GHz and an NVIDIA Titan X GPU. We test using sythetic data for which we have produced a sequence of point clouds with a large amount of non-overlap, an indoor Lidar sequence inside a factory (Velodyne VLP-16), and the KITTI outdoor Lidar sequence 04 [10]. For completeness, we also tested on a Kinect dataset (Stanford Lounge) [27], which we have included in our supplementary materials.

### 4.3. Robustness to Partially Overlapping Views

To measure robustness to non-overlapping regions in a controlled manner, we decided to create a synthetic dataset where there exists a challenging amount of non-overlapping geometry. To do this, we created five partially overlapping point clouds of the Stanford Bunny (Fig. 4(top row)) using a simulated and constrained 60 degree field-of-view. To compare the performance of different algorithms, we register together the point clouds, frame-by-frame, in sequence. This particular sequence represents a challenging scenario for registration algorithms since the amount of overlap present varies between frames. For example, between Frame 1 and Frame 2, the bunny's head and front half is cut off and is only visible in the latter. To register this data together properly, the head and front half points in the second point cloud must be ignored, either as outliers, or in such a way that this nonoverlapping region is downweighted during distance minimization.

To visually compare the results among different algorithms, we placed all five frames into a global coordinate system by compounding successive poses (Figure 4). The middle row of Fig. 4 shows the results of ICP [3], Trimmed ICP (TrICP) [5], MLMD [7], FICP [21], and Iteratively Reweighted Least Squares ICP (IRLS-ICP) [13].

Without EOE (Figure 4, middle row), all algorithms tested but MLMD show significant error (average rotational error along each rotation axis was > 10 degrees). These methods tend to converge to suboptimal solutions where the centroid of each point cloud are maximally overlapping, without regard to the specific shape information present in the data. As an example, the top row of Figure 5 shows this situation for FICP. The centroid matching effect is the result of false matches along the border of each point cloud's field-of-view extents (that is, along the border of $\Omega$): If any points just outside $\Omega$ fail to be deemed outliers, these erroneous border matches will "pull" the point clouds together incorrectly, further confusing the border for the next algrithm iteration. Thus, as the algorithm iterates, more points outside $\Omega$ are erroneously pulled inward, causing a cascading effect that converges to a bad local minimum. Because there is no way for the outlier detection system to differentiate between noise and previously unseen point data, especially near regions bordering the field-of-view extent, these types of partial overlap situations tend to cause registration divergence, even for algrorithms like Trimmed ICP, which are designed to be resilient to non-overlapping regions.

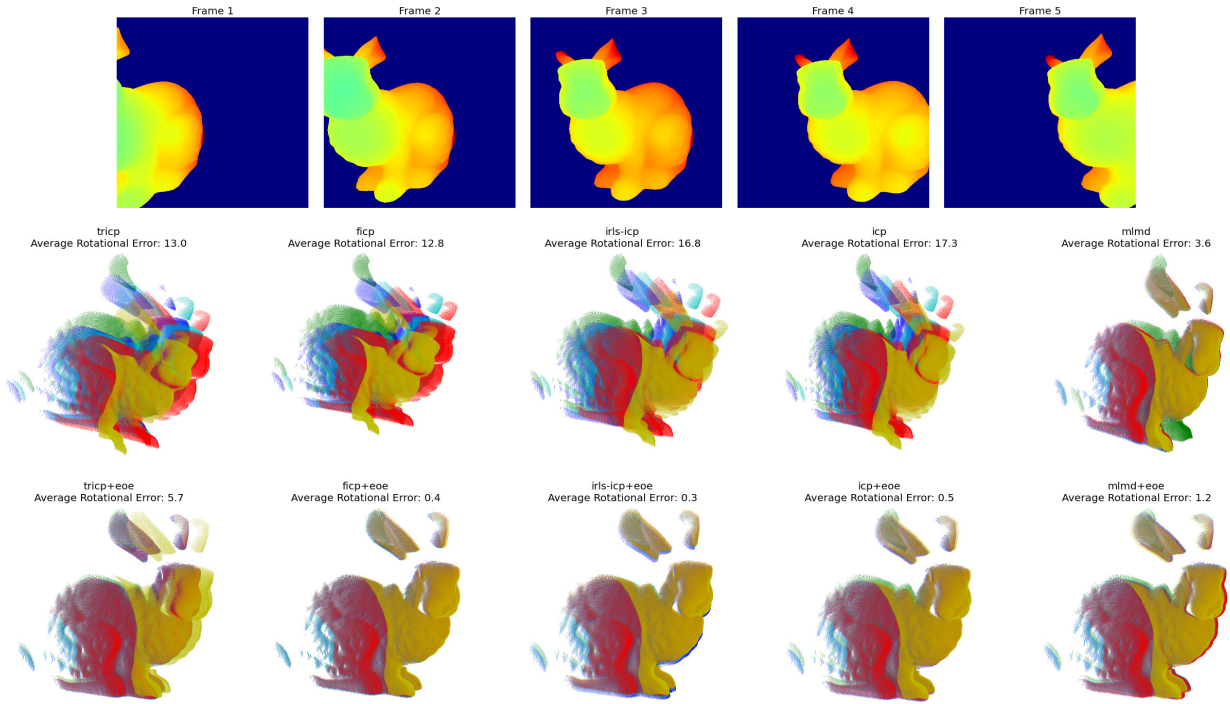

Figure 4. **Partially Overlapping Bunnies** Five partially overlapping point clouds taken from the Stanford Bunny were registered together in sequence and then placed into a global coordinate system. Each of the five views are shown with different colored points. **Top row** Input depth stream used for bunny overlap experiments. **Middle row** Registration of each algorithm without EOE. High average error produces highly mis-aligned data. **Bottom row** Registration of those same algorithms augmented with EOE. The addition of EOE improves accuracy by up to two orders of magnitude.

In contrast, the introduction of EOE (bottom row of Figure 4) drastically reduces the average error over each of the 5 registered partial bunny point clouds. For example, IRLS-ICP's average angular error reduced from 16.8 degrees without EOE, to 0.3 degrees using EOE, and Trimmed ICP reduced from 13.0 to 5.7 degrees using EOE. Refer to

Table 1. **Accuracy Comparisons on Bunny Dataset**

| Method | Avg. Error (degrees) | |
|---:|:---:|:---:|
| | Normal | with EOE |
| ICP | 17.3 | 0.5 |
| MLMD | 3.6 | 1.2 |
| FICP | 12.8 | 0.4 |
| IRLS-ICP | 16.8 | 0.3 |
| TrICP | 13.0 | 5.7 |

Table 1 for the numerical results. By iteratively estimating the sensor's field-of-view extents, we are able to statistically discount points outside $\Omega$, removing their influence and tendency to corrupt the solution, and thus registering together only the point data seen in both views. See the bottom row of Figure 5 for a graphical depiction of this process, where as the algorithm iterates, the red region (downweighted region expected to be outside $\Omega$) grows until the true solution is reached.

### 4.4. Lidar Data

Though the Velodyne VLP-16 LIDAR has a 360 degree horizontal field-of-view, it suffers from a relatively narrow vertical field-of-view of 30 degrees. Similarly, the angu-

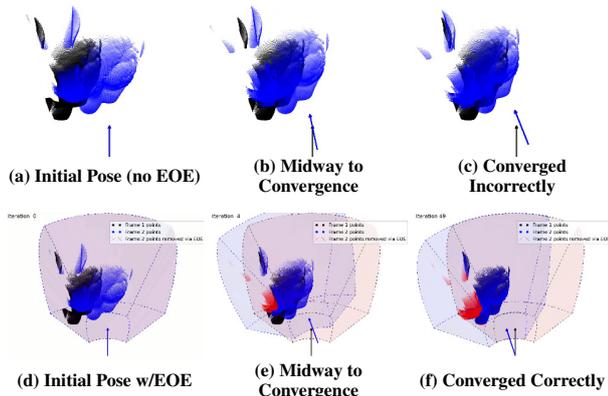

(a) Initial Pose (no EOE)  (b) Midway to Convergence  (c) Converged Incorrectly

(d) Initial Pose w/EOE  (e) Midway to Convergence  (f) Converged Correctly

Figure 5. **FICP on Bunny Dataset** Frames 1 and 2 of the Bunny dataset are shown in black and blue. *Top Row*: The progression of FICP (without EOE) from beginning state to final convergence (left to right). Note how the point cloud centroids align, causing an incorrect result. *Bottom Row*: We visually depict EOE via field-of-view constraints. As the algorithm iterates, more and more points are correctly estimated to be outside $\Omega$ (shown in red in (e) and (f)). Note that correctly down-weighting points outside $\Omega$ allows FICP to find the true minimum.

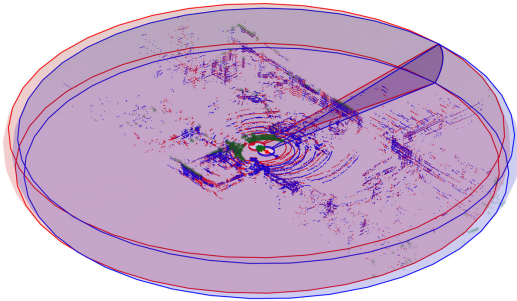

Figure 6. **Velodyne View Model** Though the Velodyne VLP-16 LIDAR has a 360 degree horizontal field-of-view, it suffers from a relatively narrow vertical field-of-view of 30 degrees. These properties lead to the formation of dense rings of sampled geometry around the sensor, which tend to be problematic for data association.

lar resolution of the pitch and yaw axes vary dramatically. These properties lead to the formation of dense rings of sampled geometry around the sensor, which often will degrade the accuracy of point-to-point matching algorithms.

### 4.4.1 Velodyne VLP-16

For a qualitative field test, we placed a Velodyne VLP-16 on a moving platform and collected a single sweep of data (roughly 0.10 seconds worth) every 3 seconds inside a large factory-like setting. Figure 6 shows a depiction of the view model used in conjunction with EOE. We registered each sweep successively into a single global frame using vari-

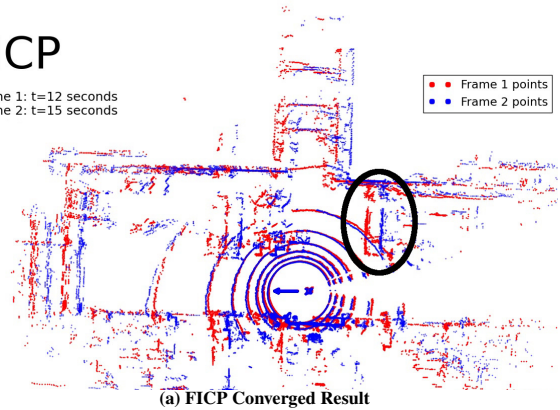

(a) FICP Converged Result

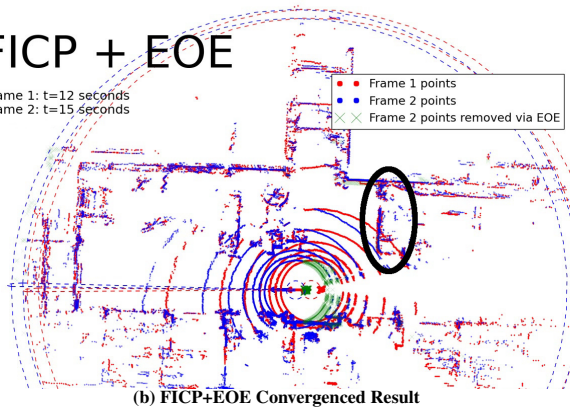

(b) FICP+EOE Convergenced Result

Figure 7. **Problematic LiDAR Registration Scenario** Two LiDAR data frames (red and blue) are registered together using FICP. This is a top-down view of the final registered output with and without EOE (points downweighted by EOE are shown in green). Note the L-shaped partition inside the black ellipse in both cases: FICP augmented with EOE produces the correct result while FICP without EOE incorrectly matches the ring-like floor points instead of the L-shaped partition.

ous algorithms. The results of these algorithms with and without EOE can be seen in Figure 8. Through visual inspection, one can see how most algorithms when not using EOE tend to underestimate pose change. To show why this is happening, refer to Figure 7, in which we inspect a pair of data frames and plot FICP's converged output in a top-down view. The problem is that FICP, even with robust outlier detection, mistakenly tries to align the circular sampling pattern on the floor instead of real geometry. Given that these rings of points are returns from a large expanse of flat ground, their appearance will not change very dramatically with any pose change on the x, y, or yaw axes. Thus, a registration algorithm that matches these ring patterns is likely to produce a "no-movement" result. However, FICP augmented with EOE produces a correct result by excising points outside the Velodyne LiDAR's narrow vertical field-of-view (points outside $\Omega$ shown in green), which is enough

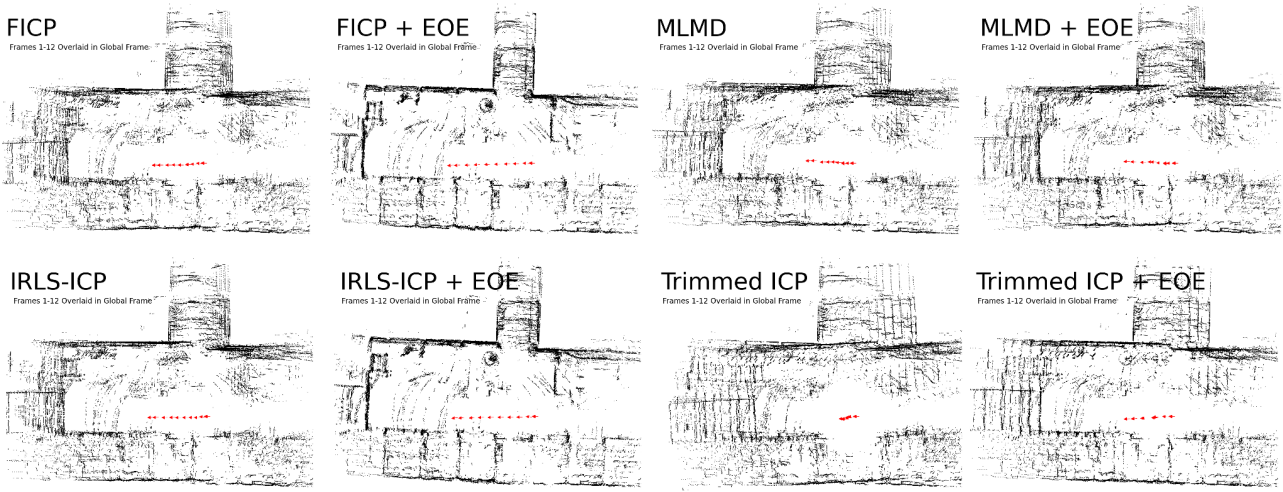

Figure 8. **Qualitative Comparison (LiDAR)**: This is a top-down view of twelve frames of Velodyne LiDAR data that were registered frame-to-frame and then plotted in a global coordinate system. The ground points have been removed for clarity. Look to the northern corridor for the clearest signs of registration error. Note the severe inadequacy of Trimmed ICP without the addition of EOE, which produces too little movement. IRLS-ICP and FICP+EOE produce the best results.

for FICP to converge to a correct solution where it leaves the floor points correctly unmatched.

### 4.4.2 KITTI Datset (Velodyne HDL-32E)

We tested against the popular KITTI dataset, sequence 04, which was taken by driving on a road in a suburban area with trees and other cars [10]. This dataset was created with a Velodyne HDL-32E, which has a very shallow vertical field-of-view (26.9 degrees). To make the sequence harder, we decrease the amount of overlap between successive frames by registering every fifth frame in the sequence and downsampling to 10,000 points (i.e. we register frame 1 to frame 5, and frame 5 to frame 10, and so on). For rotation error, we calculate the average Euler angle deviation in degrees from ground truth and the translation error is measured in meters. These measures give us a relative estimate of drift per frame with and without the application of EOE. The average and median frame-to-frame errors in rotation and translation are shown in Table 2. Across all algorithms tested, EOE reduces the amount of drift accumulated per frame, though the effect is most noticeable in the translation error compared to the rotation error. For example, for FICP the median translational error decreased from 2.37 meters per frame to 102 mm per frame. We note that the translation error decrease can be explained by the same situation seen in Figure 7.

## 5. Conclusion and Future Work

We derived a new algorithm for overlap estimation that can be added to most existing point cloud registration techniques. Our motivation was based on the observation that registration algorithms utilize assumptions that are often violated in real-world conditions, namely, that points can be thought as having been generated *iid* by some latent probability model. Thus, we developed an EM procedure to estimate the latent overlap in conjunction with the spatial model and transformation parameters. For common failure cases in unstructured domains, EOE tends to provide more accurate and robust transformation estimates and can be seen as complementary to existing outlier detection methods in the literature.

Most ICP-based and GMM-based algorithms can be augmented with EOE fairly easily and with little overhead. Given that EOE can be run in parallel per-point, we developed a fast and lightweight CUDA kernel that can scale easily to 100's of thousands of points on a modern GPU. Our implementation will be made public as an extension to existing software packages.

Table 2. **KITTI Sequence 04** Every fifth frame in Sequence 04 registered together frame-to-frame. Median and average error per frame shown with and without EOE.

| Method | Avg / Median Err (°) | | Avg / Median Err (m) | |
|---|---|---|---|---|
| | Normal | w/EOE | Normal | w/EOE |
| MLMD | .21 / .16 | .19 / .14 | 1.24 / .817 | 1.17 / .103 |
| ICP | .32 / .28 | .30 / .25 | .933 / .233 | .636 / .182 |
| FICP | .18 / .13 | .13 / .07 | 1.99 / 2.37 | 1.15 / .102 |
| IRLS-ICP | .22 / .17 | .18 / .11 | 1.29 / .683 | .924 / .109 |
| TrICP | .19 / .14 | .13 / .10 | 2.08 / 2.41 | 1.56 / 1.94 |